\documentclass[sigconf]{acmart}
\usepackage{multirow}
\usepackage{xcolor}
\usepackage{soul}
\usepackage{graphicx}
\usepackage{natbib}
\usepackage{dirtytalk}
\usepackage{url}
\usepackage{balance}
\usepackage{stfloats} 

\usepackage{xspace}
\newcommand{\syste}{fAIlureNotes} 
\newcommand{\system}{fAIlureNotes\xspace} 
\newcommand{\systems}{fAIlureNotes's\xspace} 

\makeatletter
\newcommand{\thickhline}{%
    \noalign {\ifnum 0=`}\fi \hrule height 1pt
    \futurelet \reserved@a \@xhline
}

\copyrightyear{2023} 
\acmYear{2023} 
\setcopyright{acmlicensed}\acmConference[CHI '23]{Proceedings of the 2023 CHI Conference on Human Factors in Computing Systems}{April 23--28, 2023}{Hamburg, Germany}
\acmBooktitle{Proceedings of the 2023 CHI Conference on Human Factors in Computing Systems (CHI '23), April 23--28, 2023, Hamburg, Germany}
\acmPrice{15.00}
\acmDOI{10.1145/3544548.3581242}
\acmISBN{978-1-4503-9421-5/23/04}

\AtBeginDocument{%
  \providecommand\BibTeX{{%
    \normalfont B\kern-0.5em{\scshape i\kern-0.25em b}\kern-0.8em\TeX}}}

\begin{document}


\title{\system: Supporting Designers in Understanding the Limits of AI Models for Computer Vision Tasks}
\author{Steven Moore}
 \email{steven.moore@tum.de}
\affiliation{%
  \institution{Technical University of Munich}
  \city{Munich}
  \country{Germany}
}

\author{Q. Vera Liao}
 \email{veraliao@microsoft.com}
\affiliation{%
  \institution{Microsoft Research}
  \city{Montreal}
  \country{Canada}
}

\author{Hariharan Subramonyam}
 \email{harihars@stanford.edu}
\affiliation{%
  \institution{Stanford University}
  \city{Stanford}
  \country{USA}
}

\renewcommand{\shortauthors}{Moore et al.}



\begin{teaserfigure}
  \centering
   \includegraphics[width=\textwidth]{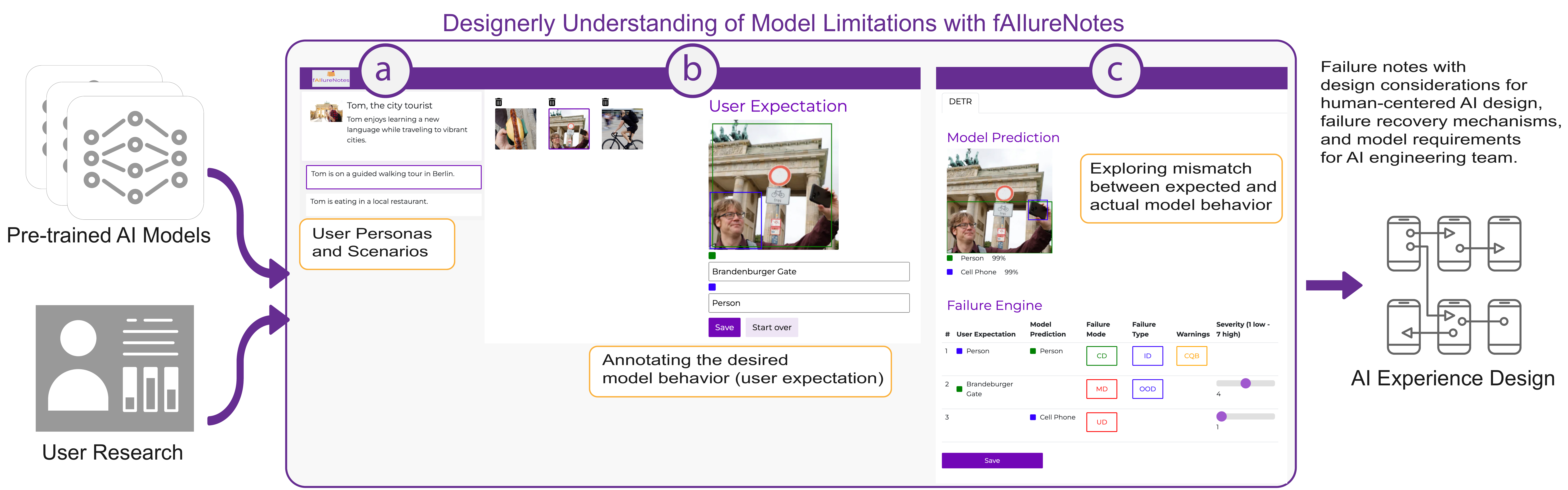}
 \caption{Designerly understanding of AI model failures using \system: (a) The designer selects a user scenario based on their user research and (b) annotates the expected behavior of the AI model for an input image. Running the model for the image, (c) they compare differences in expected and actual behavior. \system summarizes the mismatches according to different AI failure modes.}
  \label{fig:teaser}
  \Description{\system supports a designerly understanding of AI models. Pre-trained AI Models and User Research are integrated into an iterative failure exploration and analysis approach through which designers create design considerations from the AI experience.}
\end{teaserfigure}

\begin{abstract}

To design with AI models, user experience (UX) designers must assess the fit between the model and user needs. Based on user research, they need to contextualize the model's behavior and potential failures within their product-specific data instances and user scenarios. However, our formative interviews with ten UX professionals revealed that such a proactive discovery of model limitations is challenging and time-intensive. Furthermore, designers often lack technical knowledge of AI and accessible exploration tools, which challenges their understanding of model capabilities and limitations. In this work, we introduced a \textit{failure-driven design} approach to AI, a workflow that encourages designers to explore model behavior and failure patterns early in the design process. The implementation of \system, a designer-centered failure exploration and analysis tool, supports designers in evaluating models and identifying failures across diverse user groups and scenarios. Our evaluation with UX practitioners shows that \system outperforms today's interactive model cards in assessing context-specific model performance.

\end{abstract}


\maketitle
\bibliographystyle{ACM-Reference-Format}

\section{Introduction}
The probabilistic and evolving nature of AI has demanded changes in the user experience design process. When crafting the user experience (UX) for an AI-powered application, product designers need to understand the capabilities and limitations of the AI model, anticipate model breakdowns, and provide users with a path forward from failure. In characterizing AI as a ``design material'', prior work recommends that designers build their understanding of AI models by connecting models' material properties to end-user needs. Unfortunately, designers currently lack the means to \textit{explore} AI models in a way that is compatible with their needs and work practices. Moreover, designers often lack the necessary technical knowledge, and the complexity of model behavior can pose challenges in understanding AI limitations, such as discovering ``unknown unknowns''---cases in which the model outputs a wrong prediction with high confidence.

Further, UX-AI collaboration can be difficult due to a lack of common language or shared processes~\cite{yang2020}. 
In many cases, this results in an \textit{isolated} AI experience design and development (AIX) process in which engineers develop the model with a limited understanding of its precise usage context while designers conceptualize the user experience with little direct exposure to the underlying technology. \\
The UX-AI gap is becoming even larger with the popularity of off-the-shelf AI models---pre-trained models that are published on platforms such as \textit{Hugging Face}~\cite{hugging}. Such models can make a valuable resource for crafting AI UX by providing designers access to numerous models trained in a variety of AI capabilities, including object detection, image segmentation, and question-answering. However, they are often pre-trained on unknown or generic data. Documentation such as model cards provides an overview of training, performance, and intended use but can be insufficient for designers to assess the fit for their specific application contexts. More generally, AI failures and their consequences are often tied to specific user data and use contexts (e.g., failure to recognize text in a foreign language can disadvantage users who speak that language). Therefore, designers need to (1) explore pre-trained AI models by considering diverse users of their product and their contexts of use~\cite{yang2018,subramonyam2021} and (2) come up with design considerations for handling AI failures such as providing warnings or explanations, or handing over the control to the user.

Through formative need-finding interviews, we learned that, at the current time, designers and product managers rely largely on Wizard of Oz simulations to explore model behavior rather than directly interacting with the model. Proactive approaches to model understanding are uncommon, as they are considered to be resource-intensive.  Instead, designers respond reactively to model failures observed during controlled deployments such as beta launches. This approach is problematic as both upstream and downstream changes can be costly~\cite{mcgregor2020, hong2021}. Practitioners need tooling support for model exploration centered around their application and end-users, supporting the discovery of a range of AI failures that may require design interventions. To address these needs, we investigate \textit{how to support a designerly understanding of the capabilities and limitations of pre-trained AI models with a focus on computer vision tasks such as object detection in images.}


To facilitate this understanding and encourage a proactive approach to AI failures, we introduce \system, a model exploration and analysis tool. \system implements an integrated workflow for guided model exploration, failure assessment, and UX design synthesis. Starting with user groups or personas, the designer can explore how the model behaves in a disaggregated manner with different inputs. For example, imagine designing a photo-based language learning application, one that identifies objects in images in a new language. Ideally, such an application should work in a variety of contexts, such as a tourist walking through the city center in Germany or an immigrant exploring items in a supermarket. As shown in Figure~\ref{fig:teaser}, the designer selects an input (in this case, an image for language learning) and annotates it from a user's perspective. Calling the model API returns the model output for the input image, highlighting any mismatch between the annotated end-user expectations and the actual behavior. Furthermore, \system adopts a novel data augmentation approach to support exploring alternative inputs by leveraging an image generation model. Through iterative exploration, the designer can develop an in-depth understanding of the model behavior, examine the performance of the model for different personas, and synthesize design considerations based on observed model failures. 

\system provides a designerly way to explore computer vision models with image data  by incorporating outputs from user research (e.g., common user scenarios and data instances) into the model exploration process. Our interactive interface and scaffolding features enable designers to understand different types of model errors and proactively identify design interventions for AI failure cases. \system also enables a necessary first step in working with pre-trained AI models as a design material. Our key contributions include (1) design considerations for tools supporting designers' model exploration derived from need-finding interviews with practitioners, (2) \system --- a model behavior analysis tool with an automated failure engine that is based on a taxonomy of failure modes for computer vision tasks, and (3) empirical evaluation of our model exploration workflow.


\section{Related Work}
We situate our research within prior literature on AI design challenges, tools for AI model behavior analysis, and existing approaches for AI failure exploration and recovery mechanisms.

\subsection{AI Design Challenges}
Prior studies have highlighted different types of challenges in designing and prototyping AI UX. \citeauthor{yang2020} mapped out numerous challenges designers face in the user-centered design process (double-diamond). For example, designers have difficulty articulating what AI can and cannot do, sketching divergent AI interactions, or anticipating unpredictable AI behaviors~\cite{yang2020}. The probabilistic and evolving nature of AI systems makes it difficult for designers to understand AI as design material~\cite{yang2020}, while designers' limited technical understanding may prevent them from understanding the capabilities of the AI model~\cite{dove2017, yang2018}. Designers often need to work closely with engineers to learn more about the technology and its capabilities and limitations. Still, such collaboration can be difficult because there is often a lack of common language or shared processes~\cite{kayacik2019, yildirim2022, windl2022}. Breakdowns in interdisciplinary communications can prevent efficient UX-AI collaboration and slow down the AI development process~\cite{hong2021, yang2018, subramonyam2022a}.

During prototyping, designers struggle to capture and test the dynamic behaviors of AI products. To reduce the technical development overhead, designers often turn to Wizard of Oz (WoZ) methods~\cite{browne2019, cranshaw2017, klemmer2000, begel2020, subramonyam2021towards}. However, WoZ can be overly optimistic and easily overlook important details of AI implementation and the complexity of AI's output space, or fail to achieve a realistic error representation~\cite{begel2020,jansen2022}. With these difficulties, AI product teams often skip early prototype testing entirely or overly focus on ideal user journeys overlooking possible AI failure cases that deviate from this ``golden path''~\cite{hong2021}. Generally, designers lack tools that support their design and prototyping processes for AI-powered products. \system aims to enable designers to have exposure to the underlying technology in the early design phases. Through iterative model probing, designers can realistically capture the dynamic AI behavior and derive design considerations for cases where the system fails.

\subsection{Behavioral Analysis of AI}
Analyzing AI model behavior and performing error analysis can reveal more nuanced error patterns not captured by aggregate performance metrics. When performing model behavior analysis, AI engineers typically collect inputs from a variety of sources, such as users or synthetic data collection, and examine their model outputs. Engineers must then organize the inputs or outputs into schemas of semantically similar samples. Schemas can be groups, clusters, or slices of data. A common example of such a schema for model outputs is confusion matrices for classification problems~\cite{oakden-rayner2020}.  Many other model-related problems can be found by grouping the inputs, often referred to as ``data slicing'' or ``subgroup analysis.'' The analysis of model performance across subgroups forms the basis for much of the work assessing the fairness of AI systems~\cite{koenecke2020, obermeyer2019}. In particular, disaggregated evaluation~\cite{solon2021}, by comparing performance metrics across sub-groups, can help uncover biased systems, such as gender classification models that significantly underperform for women of color~\cite{buolamwini2018}. Testing AI iteratively with different inputs helps engineers formulate hypotheses about how and why the model fails and how to improve the model. The final evaluation of a model behavior analysis is usually documented in the form of reports such as a checklist \cite{madaio2020} or model card~\cite{mitchell2019, anamaria2022}. These reports can include performance metrics and major failure modes to facilitate suitability assessment and responsible use for those who want to use the model. 

Several tools have been developed to help engineers analyze the behavior of AI models from input generation \cite{ratner2017, cashman2020, wexler2020}, schema construction~\cite{polyzotis2019, nushi2018, amershi2015, bauerle2022}, hypothesis definition~\cite{wu2019, eyuboglu2022}, to final evaluation~\cite{arnold2019}. For example, Affinity~\cite{cabrera2022} allows AI engineers to filter images and visually inspect the model output. The tool also enables engineers to group images into schemas and define group-specific hypotheses as to why the model failed.  Beat the Machine~\cite{attenberg2015}, DynaBench~\cite{kiela2021}, and Patterned Beat the Machine~\cite{liu2020} support input data collection by encouraging end users to find instances where the model fails. Another method to collect challenging data samples is data augmentation, by modifying existing instances to create new instances, e.g., by rotating or cropping images~\cite{suorong2022}, or using recent AI techniques to generate new artifacts~\cite{figueira2022} such as generative adversarial networks (GANs)~\cite{goodfellow2014, karras2018}.

All of the above tools aim to support engineers; designers have received little attention. \system is a model behavior analysis tool built for designers (broadly defined) rather than machine learning engineers. Similar to Beat the Machine~\cite{attenberg2015}, we encourage designers to find images that break the model. Similar to Affinity~\cite{cabrera2022}, our tool allows designers to group error cases into categories prior to assigning appropriate design interventions as recovery strategies. To allow data augmentation, \system  uniquely integrates a generative text-to-image model (diffusion models) \cite{stabilityai}. It supports the creation of prompts (text strings) describing desired images and feeds the prompts into the generative model to obtain the image instances. This approach solves a critical problem that product- and context-specific data can be costly to collect, especially in the early stage of product development.  To support non-technical UX practitioners, \system also allows the creation of user scenarios to ground the exploration of model inputs and outputs, and contains a failure engine that automatically identifies and explains different failure modes to designers. Lastly, \system also allows designers to evaluate how well a model fits their various user groups by reviewing disaggregated failure metrics, a concept that was inspired by disaggregated evaluation~\cite{solon2021}.

\subsection{AI Failure Exploration and Recovery}
Numerous guidelines have been proposed to design AI experiences (AIX)~\cite{ibm2019, apple2019, hagendorff2020, jobin2019}, including recommendations to mitigate and recover from AI failures~\cite{pair2019, amershi2019}. The PAIR playbook, for example, dedicates an entire chapter to errors and ``graceful failure'', offering best practices for designers to identify and diagnose AI errors and contextual failures while also providing potential paths forward from failures~\cite{pair2019}. Microsoft's guidelines for human-AI interaction advise designers to make clear what the system can do and how well it can initially perform those tasks. If the system fails, designers should support efficient correction, scale down AI influence when in doubt, or provide users with local explanations~\cite{amershi2019}. Horvitz's principle of mixed-initiative user interface design~\cite{horvitz1999} provides another best practice when designers decide to turn control over to humans in the face of AI uncertainties. These recommendations often provide high-level guidance but lack actionable details.

Wizard of Error~\cite{jansen2022} allows designers to perform Wizard of Oz (WoZ) studies and simulate ML errors. To envision possible failures early in the design process, \citeauthor{hong2021} developed the HAX Playbook, a low-cost tool that promotes proactive consideration of model-related failures in natural language processing. Neither tool allows working with a specific AI model, making it difficult for designers to get a sense of the realistic model behavior. Gradio~\cite{abid2019} allows AI engineers to share their developed ML model with non-technical collaborators or end users. The Python package generates a visual interface that can be used to probe the model, view model outputs, and flag suspicious predictions. Alternatively, ProtoAI~\cite{subramonyam2021} uses a model-based prototyping approach that allows designers to incorporate model outputs directly into UI designs and analyze model failures as they occur. \system complements these approaches and helps designers find AI-related failures before deployment. Unlike ProtoAI, we place a stronger focus on discovering AI failures while developing a material understanding (i.e., before the design phase). 
\section{Formative Interviews}\label{sec:probe}
Our goal for the formative need-finding interviews was to (1) investigate current practices of UX designers to understand the limitations of AI models, and (2) derive design considerations for our tool by observing practitioners probing existing model APIs for errors. 
The formative study does not aim for a comprehensive sample but is intended to determine system requirements based on qualitative feedback from a qualified and skilled group of participants.

\subsection{Method}
\subsubsection{Procedure} 
We conducted semi-structured interviews with ten UX practitioners (UX researchers, product designers, and product managers) recruited through social media and using snowball sampling. To assess fit, potential candidates were asked to fill out a short form before participation. All participants had prior experience working with AI-enabled products with an average experience of 5.2 years ($SD=2.4$ years) (see Table~\ref{table:demography}). 

\begin{table}[h!]
\begin{tabular}{ l l l }
\toprule
\textbf{ID} & \textbf{Role} & \textbf{Work Experience in AI} \\
\midrule
P1 & UX Researcher & 10 \\
\hline
P2 & UX Researcher & 4 \\
\hline
P3 & Senior UX Researcher & 1 \\
\hline
P4 & Lead Product Designer & 6 \\
\hline
P5 & Product Manager & 5 \\
\hline
P6 & Product Manager & 8 \\
\hline
P7 & Senior UX Researcher & 4 \\
\hline
P8 & Product Manager AI for Design & 4 \\
\hline
P9 & UX Researcher & 6 \\
\hline
P10 & AI Product Manager & 4 \\
\bottomrule 
\end{tabular}
\caption{Demography of Study Participants}
\label{table:demography}
\Description{Table showing the demography of study participants including an ID, participants' role and AI work experience.}
\end{table}

Each interview was conducted using the Zoom video conference tool and lasted approximately 60 minutes (a total of 10 interviews). Participants received a \$30 Amazon gift card as an appreciation for their time. The study was approved by our Institutional Review Board.
In each interview, the first 30-40 minutes were used to explore existing approaches to understand model-related failures. Using an application context from participants' own work as an anchor, we questioned them about AI failure discovery and understanding, collaboration, knowledge sharing with engineers, and design approaches to address AI failures in their products. We also asked participants about specific methods, processes, and tools they use to learn about the capabilities and limitations of AI models. In the last 20-30 minutes of the interview, participants engaged in a hands-on activity to explore a pre-trained model using a provided application scenario (similar to~\cite{subramonyam2021}). We intended to understand whether and how model documentation and interactive APIs facilitate a designerly understanding of model limitations. 

We asked participants to imagine designing an AI-powered language learning application that helps beginners learn single words in a foreign language. Specifically, end-users take a picture with their smartphone camera and pass it to an object detector. The computer vision system automatically detects all objects in a scene and shows or reads the words for these objects in a foreign language.  We narrowed down on computer vision and object detection because computer vision models are popular among platforms that provide pre-trained model services. Also, image data is easy to work with, and the task is familiar to participants, as visual-verbal learning is commonly practiced in vocabulary learning. With this task in mind, we asked participants to explore Hugging Face’s pre-trained DETR model~\cite{detrhugging} to understand its capabilities and limitations. Hugging Face is a rapidly growing service that requires AI developers to produce model and dataset documentation when submitting a pre-trained model~\cite{hugging}. The documentation (i.e., model card) for DETR model includes details about the architecture and intended uses, as well as training procedure and evaluation results. Hugging Face also provided a GUI that allows model consumers to test the model by uploading images. During exploration, we encouraged participants to find challenging but realistic samples to ``break'' the model. The participants hypothesized problematic images (e.g., sketch illustration of a tomato) and used Google image search to find and test images using the interactive API.  Depending on the participants’ speed, they tested between three and ten images during the exercise. The insights from this activity helped us to derive design guidelines for \system.

\subsubsection{Analysis}
We transcribed the interview recording using a third-party service (approximately 600 minutes of recording). Using an open-coding approach~\cite{denzin2011sage} the first author coded three transcripts to create an initial codebook. The resulting codebook consists of 86 codes, including challenges in hypothesizing about errors, awareness of AI failures, perception of the severity of failures, biases in exploration, level of abstraction in documentation, and mental model building. Based on discussions with all authors, the first author revised and coded the remaining transcripts. Using MURAL, all authors grouped and categorized the codes and notes in multiple sessions and discussed the thematic relevance to the topic of successful AI failure discovery. In several cycles, we identified the most important emergent themes. We refined the themes until all categories and subtopics were covered and no new topics emerged. We used memos to summarize findings across transcripts and as a basis for discussion among the authors~\cite{birks2008}.

\subsection{Findings}
\begin{figure*}[t!]
\centering
\includegraphics[width=\textwidth]{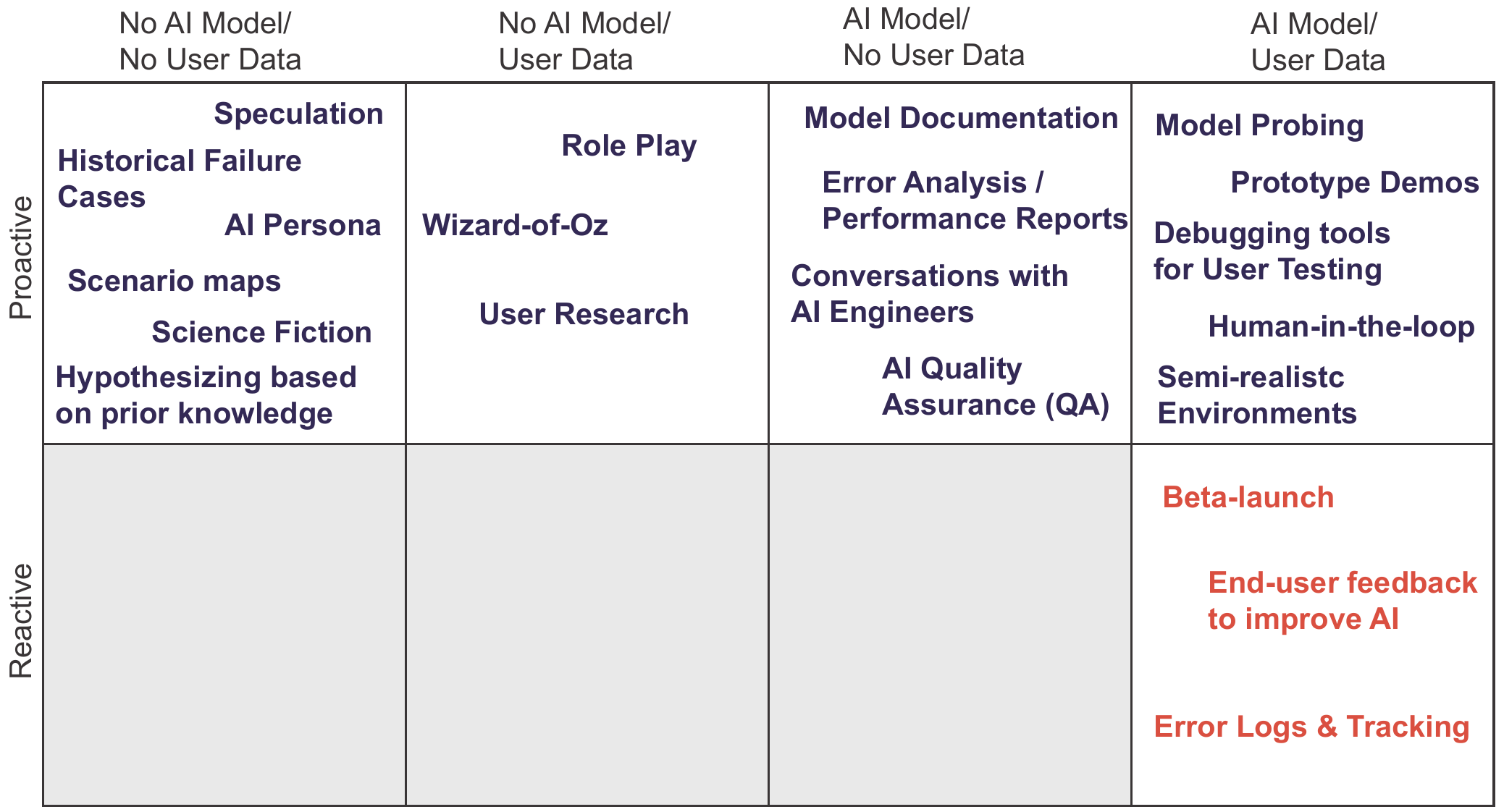}
\caption{Proactive and reactive approaches to designerly understanding of model limitations that participants discussed in the formative interview}
\Description{Figure representing designers' failure exploration methodologies classified into proactive and reactive.}
\label{figure:overview}
\end{figure*}

As summarized in Figure~\ref{figure:overview}, participants reported a variety of proactive and reactive failure exploration and model understanding strategies. We found that an effective approach to detect AI model failures is by detecting misalignment between expectations of real-world users and model behavior early on in the design ideation process. However, such proactive failure exploration is rarely practiced due to a variety of challenges, including time and effect cost, knowledge barriers, a lack of tooling support, etc. Instead, several participants relied on reactive approaches in which the first version of the AI-enabled solution is rolled out relatively quickly. End-user feedback and error logs are then used to iteratively improve the user experience and model (fine-tuning, re-training, online training). Such reactive approaches have drawbacks, including technology blindness at design time, which may create technical or design debt, and causing harms to users. Here we elaborate on \textit{proactive} approaches and challenges for UX practitioners to detect and avoid AI failures and derive design considerations for \system.

\paragraph{Proactive practices without direct access to the AI model result in an incomplete understanding of model limitations}
We found that a majority of proactive approaches today don't include the AI model itself during exploration. In the early stages of design ideation, practitioners often relied on their imagination to anticipate AI failures ex-ante (no AI Model and no User Data). Several participants made use of anecdotal evidence (e.g., science news reports) and secondary sources to understand potential AI errors. P4, a product designer, reported reading science fiction to help him think about AI model failures. According to P4: \textit{``maybe this sounds fun to you, but we also spend a lot of time reading science fiction. Trying to see what other people are predicting about possible AI failure\ldots''}. Other participants reported sketching high-level user journeys based on personas to brainstorm failure cases for each step in the journey.  To simulate AI behavior, practitioners relied on role-playing or Wizard of Oz techniques (No AI Model but User Data). P7, a product manager, commented: \textit{``Usually what I do is I'll partner with a designer, and we will create a high fidelity prototype that I will Wizard of Oz. So, I will pretend to be the model for the purpose of qualitative discovery''}. However, such approaches alone can be overly optimistic~\cite{begel2020}, and designers may have difficulty imagining all mistakes that an AI would make~\cite{jansen2022}.  

Additionally, as observed during hands-on exploration, with access to AI models, participants were able to refine their initial hypotheses about model limitations by observing failure outputs. For instance, P5, a product manager, wanted to test the model on people of color because he had heard that computer vision models had difficulty in this area. The model could not recognize a dark-skinned person in the top row. This prompted P5 to explore the model behavior for a person who had many facial tattoos, which was also undetected by the model. Access to pre-trained models can support a richer designerly understanding of model limitations. Thus, model exploration tools should allow designers to \textbf{incorporate existing practices typical to user research with interactive AI model APIs (D1).}  We also recommend model exploration tools should allow designers to apply an \textbf{iterative approach to testing new or slightly modified hypotheses effortlessly (D2).} 

\paragraph{Contextualizing AI models within use contexts are helpful for design ideation} Three participants described approaches that allowed them to anticipate AI errors before deployment (proactively), including building specialized model exploration tools and having end-users in-the-loop (AI Model and User Data). P1, a UX researcher reported developing scenario-driven model probing tools along with AI engineers. Specifically, in the context of optimizing shelf space in stores using computer vision, P1 describes \textit{``we created this model debugging tool\ldots and we would go into stores and take images of store shelves and look at the corrections the model would recommend and debug what the AI is getting right and wrong and try to figure out why?''}. Furthermore, P1 added that this understanding was critical to designing the user interface to guide the store agent in optimizing the shelves. 

Similarly, based on observations from the hands-on exploration, we noticed that the context and needs of the user served as a starting point for testing the model for failures. For example, P8, a Product Manager, imagined a science student during a semester abroad: \textit{``if, for example, I was to study abroad in Berlin and I was doing a physics course and I wanted to know what a particular object was called, then translating would be really useful''}. She then looked for a picture of a solar system as input to the object detection model and observed that the model incorrectly classified the ``planets'' as ``balls''. Reflecting on their exploration behavior, P9, a user researcher explained that \textit{``As UX researchers, we try to ask first whom we are building it for, who's going to use it. So, that might help you to narrow down: Is there any model built for that specific type of user? Are we building it for a child or the elderly?\ldots And we then select the model based on that\ldots''} Based on this insight, we propose \textit{end-user-centricity} in model exploration tools for UX, i.e., \textbf{model exploration tools should guide UX practitioners to test the model for different user groups and contexts (D3)}. 

\paragraph{Limited AI knowledge challenges designers' understanding of model failure types and ways to handle them} Testing a pre-trained model for failures was an unfamiliar task for many of our participants. Practitioners recognized the importance of AI engineering practices (AI Model but no User Data), such as conducting technical error analysis, selecting appropriate value functions, defining performance metrics, etc. However, they reported difficulties in translating objective performance metrics into the designerly understanding of model behavior. For example, in the context of supervised learning, a model error is defined as a discrepancy between the prediction of the AI model and the ground truth. Since UX experts are concerned about UX failures that can negatively impact users, the term error is defined more broadly to include not only errors within the distribution, but also out-of-distribution and context errors. During the hands-on activity, P3, a UX researcher, imagined a tourist in France who wanted to find out about various objects in a café. She inserted a picture of a croissant, which the model incorrectly classified as a ``cat'' because it had not been trained on the class ``croissant''. The class ``croissant'' is not in the distribution of the model. P4, on the other hand, imagined that someone wanted to learn the plural. He first looked for a picture of a single apple and later for multiple apples and entered them into the model. P4 was disappointed that the model could not cover this context, even though it was a realistic case for a person who had just started to learn a new language. Without a clear view of possible error types, it is challenging for designers to ideate ways to address them in the user interface. Therefore, UX tools for model exploration should \textbf{cover a wider range of failures and support UX professionals to understand the underlying cause of AI model failure (D4).}
\section{\system}
Based on the design considerations identified in Section~\ref{sec:probe}, we implemented a prototype tool, \system, for proactive exploration of AI model failures. While our approach can be generalized to a variety of machine learning tasks, we focus our implementation on computer vision tasks using image data. Specifically, we demonstrate our approach for the task of object detection, making \system a scalable tool to probe any object detector, regardless of implementation details (e.g., \system works for single-stage \cite{girshick2013, shaoqing2015, girshick2015} and two-stage detectors \cite{redmon2015, lin2017, carion2020}). 

\subsection{User Experience of fAIlureNotes}\label{sec:ux}
\begin{figure*}[h!]
\centering
\includegraphics[width=0.7\textwidth]{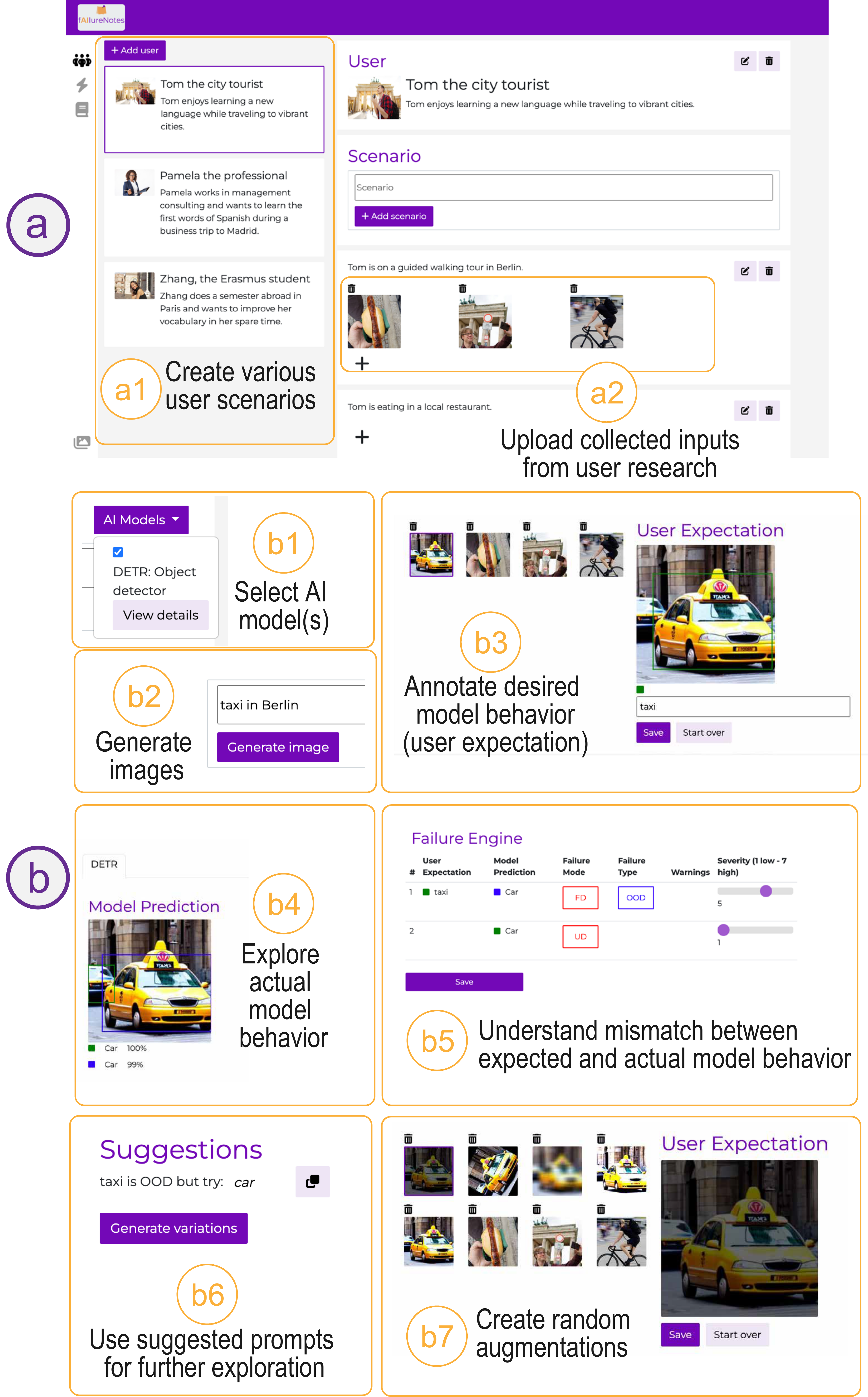}
\caption{\systems user interface for user research and model exploration: the designer (a) creates various user scenarios and uploads data inputs collected from user research. (b) The designer then conducts an iterative failure exploration.}
\Description{Figure illustrating the user experience and features of \system separated into (a) user scenarios, (b) iterative failure exploration}
\label{figure:ux_1}
\end{figure*}

\begin{figure*}[h!]
\centering
\includegraphics[width=\textwidth]{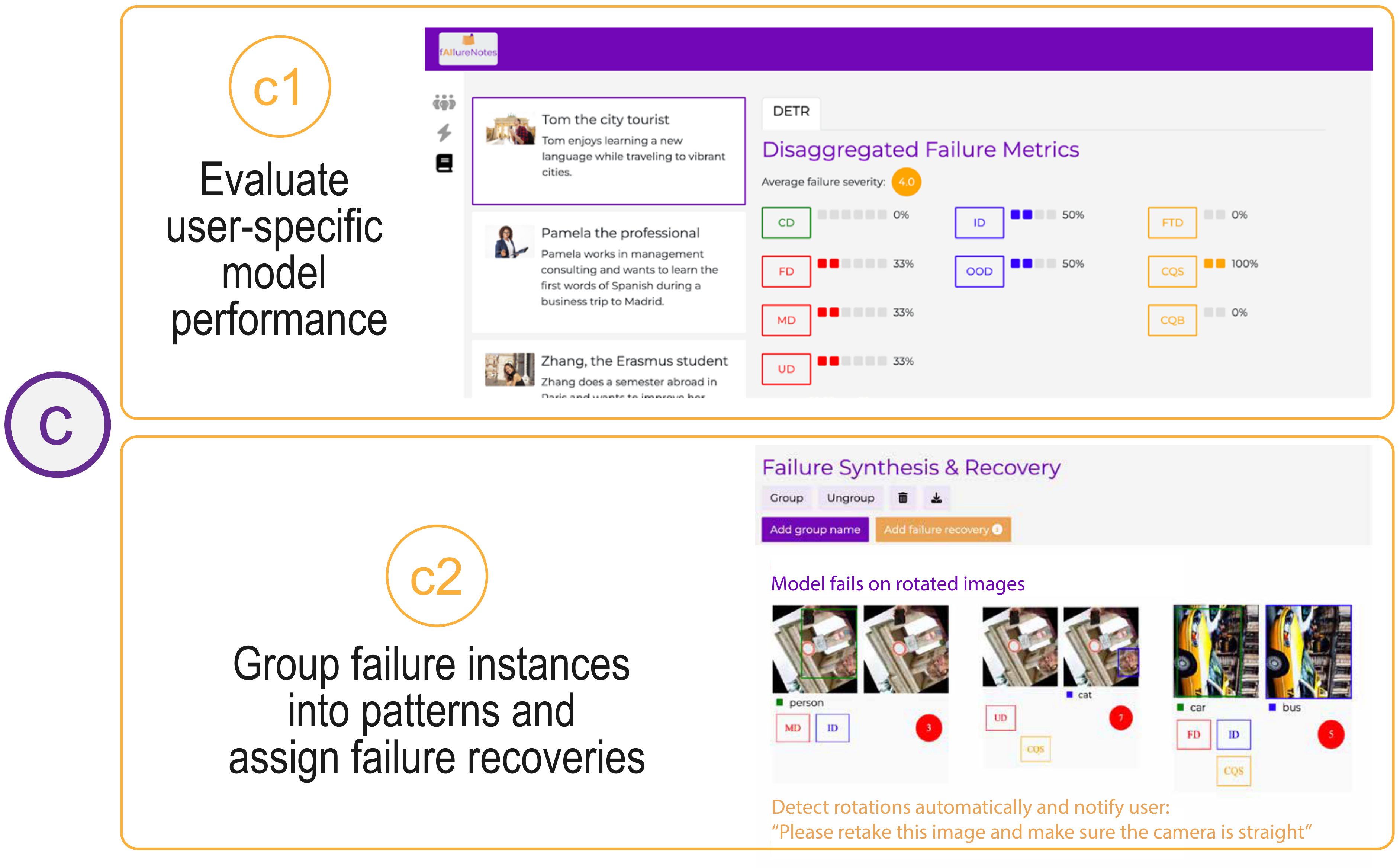}
\caption{\systems user interface for design synthesis: the designer (c) assesses model performance, reviews, and groups failure instances, and derives design considerations for recovery. }
\Description{Figure illustrating the user experience and features focusing on (c) design synthesis. }
\label{figure:ux_2}
\end{figure*}

As shown in Figures~\ref{figure:ux_1} and ~\ref{figure:ux_2}, \system consists of a tab-based layout with three main views: (1) the \textit{user scenario view} for importing insights from user research, (2) the \textit{model exploration view} for iterative model probing and failure exploration, and (3) the \textit{design synthesis view} to synthesize design considerations based on failure discovery. To better demonstrate how \system helps UX practitioners develop a designerly understanding of computer vision models and identify model failures, let us follow Eva, a UX designer with prior experience in designing AI-powered applications. Similar to the example design task used in the need-finding study, Eva wishes to design a computer vision-based foreign language learning application similar to a vocabulary trainer. 

\textbf{(a) User Scenarios:} Based on user research, Eva has identified several user groups (personas) and usage scenarios along with example data instances (i.e., images that these users would upload for language learning). Eva opens \system on her web browser and begins by importing user research data into the tool (aligned with D1, D3). By clicking on the ``add user'' button in the user scenario view, Eva creates a new persona called Tom, a tourist exploring the city of Berlin. Next, Eva adds two scenarios for Tom, including visiting the city center on a guided walking tour and eating at a local restaurant. For each scenario, she creates a brief description and uploads images typical to those scenarios by dragging and dropping them under respective scenarios (Figure~\ref{figure:ux_1}a). Eva also creates personas representing a student involved in a study abroad program, and a business professional. Note that adding images at this initial stage is not mandatory; Eva can use the \systems \textbf{inbuilt text-to-image generation functionality} to generate images by writing text prompts describing the desired image. 

\textbf{(b) Iterative Failure Exploration:} Next, Eva opens the model exploration view to understand how well object detection models work for her application design context. As shown in Figure~\ref{figure:ux_1}b, Eva has access to the user scenarios displayed on the side panel on the left. The main view shows a multi-selector dropdown on the top with a list of AI models and a prompt-based image generator right below it. Eva selects the DETR Object Detection model to begin exploring its capabilities and limitations. This creates a new tab for the DETR model in the main view. She can inspect the model documentation, if any, by clicking the ``view details'' button next to the model. 

Now Eva is ready to explore the model behaviors with different input images. Rather than choosing images at random, \system scaffolds Eva to assess the model behavior and limitations for each of the user scenarios created in the previous step. Eva selects a specific user scenario, ``Tom is on a guided walking tour'', and since Eva does not have a user-provided image for that, she generates an image of a Taxicab  with the text prompt ``Taxi in Berlin''. Two images appear and Eva selects one for further analysis by clicking on it. The selected image appears in the main model exploration view. Instead of directly running the model, Eva first annotates her expectations about the model by creating bounding boxes and labels for objects in the image (in this case a single bounding box around the Taxicab in the image and the object ``taxi''). This approach allows designers to input their definition of correct model behavior, and later analyze differences in the actual model behavior to detect failures. Next, Eva runs the DETR model for the same image and the model outputs two objects, both labeled ``car''. She notices that the object she labeled ``taxi'' is detected as a ``car'' and another ``car'' is detected in the background. In the model tab, each object and label in the model's prediction is assigned a unique color, and by hovering over the bounding boxes, Eva can examine the prediction object by object. 

To facilitate analysis of the mismatches, \system implements a \textit{failure engine }that automatically compares model outputs with the annotated input image and highlights different types of model failures (D4). In this case, the system matched Eva's object ``taxi'' with the corresponding ``car'' in the prediction. The system used a red error tag to classify this case as ``FD'' (false detection). The failure engine further explains that ``taxi'' is misclassified because it is not within the model's capabilities, indicating this with a blue info tag. In other words, the model has not been trained on data of this category. Additionally, the system classified the car in the background as ``UD,'' which stands for an unnecessary detection that seems less important to the user. Eva now has the option to rate the severity of each failure case on a 7-point Likert scale from the user's perspective (D1). She assigns a 5 (high severity) to the first failure by adjusting a slider and leaves the unnecessarily detected car at 1 (low severity). Clicking a button saves the two failure instances to the database. 

Last but not least, \system \textbf{suggests prompts} to generate alternative input images for Eva to continue her exploration (D2). For one suggestion feature, \system guides Eva to explore related images on which the model can produce correct outputs to help her understand the boundary of model failures. For example, here ``taxi'', is a lower-level sub-category for ``car'', so \system suggests trying with an instance image (or writing a prompt) about ``car'', which falls within the model's detectable categories. In other cases, if the model correctly detected an object in the previous exploration, the system takes common limitations of computer vision models as a starting point to suggest variations that challenge the model to fail (D4). For example, it may suggest Eva try images (or writing a prompt) such as ``a Taxistand'' or a ``bus station.'' Alternatively, Eva can use the \textbf{`Generate Variations'} feature in \system to augment her image of the Taxicab. \system generates variations of the image by adjusting brightness, rotation, and blurriness. These variations help Eva explore unknown unknowns and better understand model limitations.

\textbf{(c) Design Synthesis:} After testing the model performance for different personas and scenarios, Eva proceeds to the last step---design synthesis and generation of \textit{failure notes}. As shown in Figure~\ref{figure:ux_2}, for each persona, Eva sees a table summarising the model failure metrics including how many objects were correctly detected (CD) versus how many were misclassified (FD), missed (MD), or unnecessarily detected (UD). Such a disaggregated presentation of failure metrics helps Eva assess how well the model performs for each persona, which can help them identify disadvantaged user group that requires more UX interventions. Additionally, \system supports the comparison of multiple models. Eva can select other Object Detection models in the dropdown, compare performance across them, and find one that works best for a given or the majority of users or scenarios. 

Furthermore, \system provides a zoomable canvas for clustering failure instances and synthesizing design considerations for failure-specific recovery. The canvas is populated with all failure instances from the exploration stage as failure cards. Each card contains an image of the user's expectation (left), the model prediction (right), failure mode tags, and the failure severity Eva assigned during her failure exploration. Eva can group these cards, as well as ungroup and delete groups on the canvas. She can add group names and make notes about how to recover from the type of failure in each group in interactive text fields. She can get suggestions about failure recovery by clicking on the ``Add failure recovery'' button. This grouping process would scaffold her to discover patterns amongst failure instances by shared UX recovery mechanisms rather than by model error type alone. For instance, she groups a set of cards with rotated images and adds a group name: ``Model fails on rotated images.'' As a recovery, she imagines the application detecting rotated images automatically and notifying the user with ``Please retake this image and make sure the camera is straight'' on the UI. 
After synthesizing her failure groups and defining appropriate failure recoveries for all groups, Eva downloads each canvas as a JPEG image. The snapshot allows Eva to share her failure exploration and design considerations with the engineering team. Using her failure notes, the team can make technical choices from improving the given model (e.g. retraining the model on additional classes or fine-tuning the model with rotated images) to developing additional technical components required for the recovery mechanism (e.g., adding another model to detect rotated images).


\subsection{Taxonomy of Failure Modes and Recovery Mechanisms for Computer Vision Tasks}\label{theory}
To support the features described above, we synthesized from prior literature an initial taxonomy of different types of AI model failures for computer vision (CV) tasks and UX recovery strategies. Taking a UX perspective, we identify ways in which CV models may fail (\textit{failure modes}) as well as effective actions to prevent or recover from failures (\textit{failure recoveries}) before deploying the AI-powered system. Note that we only cover failures that occur in a single human-AI interaction (local failures) and exclude the ones that span all or several interactions, such as mental model breakdowns (global failures).  Through this synthesis, we implement the failure engine for the current instantiation of \system (see Section~\ref{implementation_details_of_tool}). 

\subsubsection{Failure Modes in Computer Vision}


\begin{table*}[h]
\begin{tabular}{ p{2cm} p{3cm} p{5cm} p{6cm}  }
\toprule
\textbf{System level} & \multicolumn{2}{c}{\textbf{Failure Mode}} & \textbf{Example Autonomous Driving}  \\
\midrule
\multirow{7}{*}{\shortstack[l]{Sensing \\ (Input level)}} & Missed or delayed attention & System fails to detect a triggering event or responds too late to be useful.
 & Motion detection system fails to trigger camera or triggers camera system too late.
\\
\cline{2-4}
& False input & System fails because it is unable to process a particular input. & System is fed lidar data but can only process camera data. \\
\cline{2-4}
& Critical quality input & System fails because of quality issues with respect to the input. & System fails because it was fed a blurry image. \\
\hline

\multirow{13}{*}{\shortstack[l]{Observation \\ (Computer \\ Vision \\ level)}} & False observation &  System generates output error.
  & Vision system misclassifies a pedestrian as another car.
  \\
\cline{2-4}
& Failing to observe & System is unable to provide an output. & Vision system is unable to detect any object for a given frame.  \\
\cline{2-4}
& Incomplete observation & System misses certain visual information.
 & Vision system misses a pedestrian in a frame.
 \\
\cline{2-4}
& Critical quality output & The output of the system bears unacceptable uncertainty. & The certainty score for the vision system’s output is too low.  \\
\cline{2-4}
& Violation & Computer vision system’s output stands in conflict with ethical standards, rules, or regulations.  & System outputs include racial categories when such information is inappropriate.
 \\
\hline

\multirow{18}{*}{\shortstack[l]{Reaction \\ (Action level)}} & Failing to act & System fails in executing the desired action. & Vehicle does not break and hits a pedestrian.  \\
\cline{2-4}
& Mistimed action & System acts correctly but at the wrong point in time. &Vehicle breaks not fast enough to prevent an accident.
 \\
\cline{2-4}
& Too much AI & From a human perspective, the system enters the territory of the user or automates too much of the process.
 & Autonomous driving assistant takes over but the human driver would prefer driving in this given situation. \\
\cline{2-4}
& Limited AI & From a human perspective, the system provides too little involvement in the process to provide value.  
 & Autonomous driving system supports human drivers with steering and acceleration but
the driver expects it also to monitor the environment.
  \\
 \cline{2-4}
& Inappropriate action & System works as intended but the behavior of the system stands in conflict with the needs, goals, or preferences of its user in a given context.
 & Autonomous vehicle speeds to be able to cross a traffic light in time but fails to take the passenger’s driving anxiety into consideration.
 \\

\bottomrule
\end{tabular}
\caption{A Taxonomy of failures in computer-vision-based systems: Describes failure modes on different system levels with examples. \system currently implements an failure engine focusing on the Observation layer}
\label{table:failure_modes}
\Description{Table of failure modes specific to computer vision technology. Failure modes are separated into three system levels: sensing (input level), observation (computer vision level), and reaction (action level).}
\end{table*}


To develop a taxonomy of typical failure modes that occur in computer vision systems, we first reviewed existing taxonomies for the field of natural language processing~\cite{hong2021} and Human-AI co-creative systems~\cite{buschek2021}. Similar to~\cite{bahaei2019}, we collected an initial set of failure modes by reviewing earlier work on human failures from the field of psychology and cognitive science. Our review included human failure types from Norman's taxonomy~\cite{norman1980, norman1981}, Rasmussen's taxonomy~\cite{rasmussen1982}, Reason's taxonomy~\cite{reason2016} as well as the HFACS taxonomy~\cite{shapell2000} which is based on Reason. We also integrated more recent failure categorizations, such as a taxonomy for human driving~\cite{stanton2009}. 

Through this review, the first author generated an initial list of failure modes. We then discussed and selected the ones that are relevant to computer vision technology. Through several meetings and discussions, we developed the categorization presented in Table~\ref{table:failure_modes}. At a high level, our taxonomy consists of three levels of failure models: input level, observation or CV model level, and reaction or response level failures. This categorization roughly follows typical CV application architecture. We list several potential failure modes for each level and also provide an illustrative example for each focusing on the use case of autonomous driving. It is important to mention that not all failure modes will apply to every vision-empowered product. Instead, the taxonomy is designed to alert a designer or engineer to a wide range of potential failures but requires the designer to select and interpret the different failure modes for their specific context. We demonstrate how this can be done with our system---\systems failure engine is based on the failure modes of our taxonomy, focusing on the observation (computer vision level) layer.
Lastly, to check for coverage and validate our taxonomy, we categorized a range of computer vision tasks using the items in our taxonomy. More specifically, we looked at the tasks of binary image classification, multi-class classification, object detection, single-object tracking, multiple-object tracking, semantic segmentation, instance segmentation, panoptic segmentation, video object segmentation, deep metric learning, and image generation. This exercise led to some adjustments in our final definitions.

\subsubsection{Failure Recovery}\label{failure_recovery}
To support the creation of failure notes, we identified a set of UX recovery mechanisms for AI failures based on human-AI guidelines (referred to as ``graceful failure~\cite{pair2019}''). We analyzed three different guidelines on AI experience design (Google’s People + AI guidebook~\cite{pair2019}, Microsoft’s Guidelines on Human-AI Interaction~\cite{amershi2019}, and Apple’s Human Interface Guidelines for Machine Learning~\cite{apple2019}). In an initial pass, the first author captured every failure recovery guideline, then we iteratively grouped similar items to identify eight commonly applied recovery mechanisms that we present in Table~\ref{table:failure_recovery}. We acknowledge that there are many more ways to prevent or recover from AI failures (e.g., engineering best practices), but the mechanisms we identified provide a starting point for design synthesis using \system. 

\begin{table*}[h]
\begin{tabular}{ p{3cm} p{10cm} }

\toprule
\textbf{Name} & \textbf{Description} \\
\midrule

\multirow{2}{*}{Quality of output} & Communicate the quality of output (e.g. confidence or uncertainty score of the prediction) to the user, or adapt user experience when confidence is low. \\
\hline
\multirow{2}{*}{N-best options} & Show the top N predictions to the user (as opposed to only presenting the prediction with the highest confidence score).  \\
\hline
\multirow{2}{*}{Hand-over of control} & Return control over to the user in situations of possible failures. Degrade the AI system’s automation level when uncertain about the user’s satisfaction.
 \\
\hline
\multirow{2}{*}{Implicit feedback} & Use implicit feedback information (e.g., users' engagement level on different outputs) to improve the AI model to align with user expectations or preferences.  \\
\hline
\multirow{1}{*}{Explicit feedback} & Consider eliciting explicit feedback to improve AI-powered experience. \\
\hline
\multirow{2}{*}{Corrections by the user} & Give users a familiar and easy way to make corrections to the AI’s output. Learn from corrections.  \\
\hline
\multirow{2}{*}{Local explanation} & Provide a local explanation for the model’s prediction. Make clear why the system did what it did. Tie explanations to possible user actions.  \\
\hline
\multirow{3}{*}{Global explanation} & Provide global explanations of how the model works. Establish appropriate trust and expectations from the beginning by communicating the product’s capabilities and limitations clearly. \\

\bottomrule
\end{tabular}
\caption{Failure recovery mechanisms extracted from Human-AI guidelines. They are provided as suggestions on the Failure Synthesis \& Recovery panel of \system.}
\label{table:failure_recovery}
\Description{Table of eight failure recovery strategies extracted from Human-AI guidelines including a name and description for each.}
\end{table*}

\subsection{Implementation Details}
\label{implementation_details_of_tool}
\begin{figure*}[h]
\centering
\includegraphics[width=\textwidth]{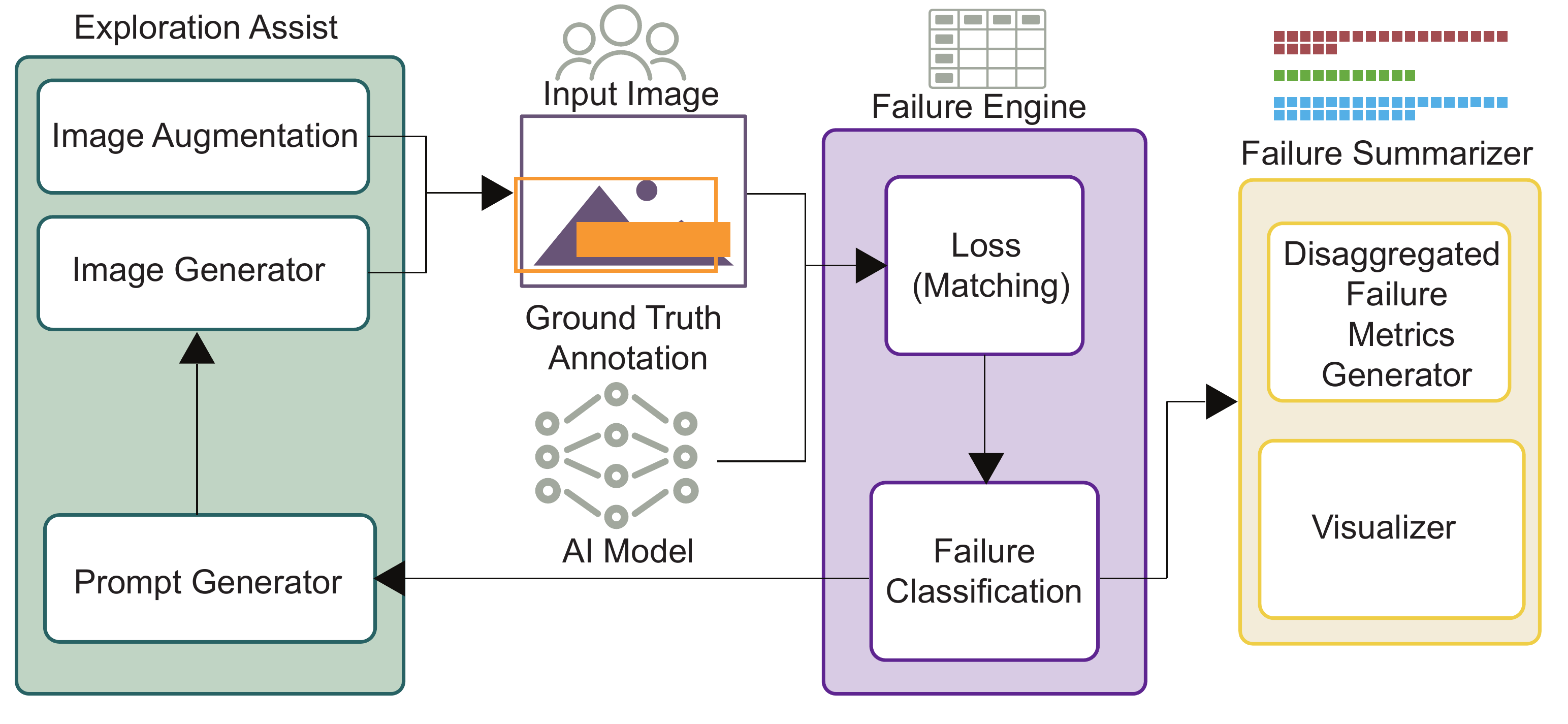}
\caption{\systems System Architecture: Input Images are annotated and fed to the AI Model. The Failure Engine calculates a Loss and performs a Failure Classification. The Exploration Assist module supports the exploration iterations and the Failure Summarizer helps assess model performance and analyze failure instances. }
\Description{Figure showing the system architecture with modules and sub-modules. }
\label{figure:implementation}
\end{figure*}

We implemented \system as a web-based application with 
Django~\cite{django}. The graphical user interface is written in HTML and JavaScript and connects to a Python Web Server that implements the modules described in Figure~\ref{figure:implementation}. 
At a high level, \textit{Input Images} can be labeled with the \textit{Ground Truth Annotation} module before being fed to the \textit{AI Model}. The \textit{Failure Engine} module calculates a \textit{Loss} and performs a bipartite matching between the annotated labels and model prediction. Based on the matches it performs an automated \textit{Failure Classification}.
The \textit{Prompt Generator} uses the information obtained from the failure engine and generates text prompts that can be fed to the \textit{Image Generator} to generate new images. Alternatively, variations of an image can be created with the \textit{Image Augmentation} module.
Lastly, the system calculates and presents \textit{Disaggregated Failure Metrics}, and provides a \textit{Visualizer} to present the failure cards in the canvas for grouping and annotating design considerations.

\paragraph{Ground Truth Annotation \& AI Model}
Designers can draw bounding boxes on an image and label individual objects. We built the annotation tool ourselves in JavaScript and did not rely on external libraries. Class labels and bounding boxes are saved in the database. For the evaluation study, we tested a common object detection network as our AI model--- the DETR model with a ResNet50 backbone \cite{carion2020}. We rely on HuggingFace's implementation support; the hosted Inference API returns the model's prediction in JSON format. A prediction contains a certain number of objects, each containing a class label, a bounding box, and a confidence score.

\paragraph{Failure Engine}
The Failure Engine conducts an automated failure classification. It takes M annotated objects (user expectation) and N predicted objects (model prediction) and creates an optimal matching based on the Hungarian algorithm~\cite{stewart2015}. The algorithm takes a matching cost as input and outputs the optimal assignment between user expectation and model prediction. Some objects may be matched and some may be left unmatched. The matching cost takes the object's class label and the bounding box into account. For the class loss, we set the cost to zero if the label of the user expectation equals the model prediction and one otherwise. The box loss is a linear combination of a simple \textit{l1} loss and the generalized Intersection over Union (IoU) loss \cite{rezatofighi2019}. The final matching cost fed to the Hungarian algorithm is the class loss added to the box loss, both weighted by a hyperparameter. We set both hyperparameters to 0.5 (identified through manual testing). \\
The matching allows us to classify different failure modes (Failure Classification) with simple rules, focusing on the observation layer of our failure mode taxonomy (Table~\ref{table:failure_modes}). If objects were matched we compare the labels between the user expectation and model prediction. If they are the same, we classify the match as ``CD'', a correct detection, otherwise as ``FD'', a false detection. Unmatched user expectations (annotated objects) are ``MD'', missing detections while unmatched model predictions are ``UD'', unnecessary detections.

The Failure Engine also checks for three kinds of warnings. If an AI model was unable to detect any object the system categorizes this as an ``FTD,'' failing to detect. If the confidence score of the prediction was below a threshold (e.g., 0.95 in our case) a ``CQS'' (critical quality score)  warning is created. In case objects have been matched and the intersection over Union (IoU) between the boxes falls below a certain threshold set to 0.7 (according to \cite{carion2020}), the warning ``CQB'' (critical quality box) is created.
Lastly, for all annotated objects (user expectation), the Failure Engine module checks whether the desired class label is in-distribution (``ID'') or out-of-distribution (``OOD''). We assume the AI's distribution is known and provided when a model is uploaded.  Interested readers can find the mathematical equations in the appendix. We relied on PyTorch \cite{pytorch}, NumPy \cite{numpy} and SciPy \cite{scipy} for the Failure Engine module. The Failure Classification is based on our failure modes (see section \ref{theory}), while the technical implementation is inspired by the object detection loss from the original DETR paper~\cite{carion2020}. 

\paragraph{Exploration Assist}
Based on information from the Failure Engine, \systems Prompt Generator proposes text prompts that can be fed into the Image Generator to generate new images during the iterative failure exploration. We distinguish three cases: \\
(1) Guide: If a tested object is out-of-distribution, we use the Words API~\cite{wordsapi} to check whether any higher-level or lower-level abstractions lie within the model's distribution. Hence, we guide designers back to the model's capabilities and help them understand the boundary of failure cases. \\
(2) Challenge: If the model predicted an object correctly, we encourage the designer to challenge the model. We create text prompts in a rule-based manner based on common limitations of computer vision models. For example, assuming the model correctly detected a cat, one of the suggested prompts may be ``An image of a cat at night'', or ``Many cats'' given that computer vision models tend to perform worse on dark or cluttered images. \\
(3) Repeat: If the model made a wrong prediction (false detection) and the annotated label was in-distribution, we encourage the designer to find a similar object as the one depicted in the image. For that, we crop the respective object and feed it to a separate image-to-text model \cite{imagetotext} to retrieve a new text prompt suggestion. \\
A suggested or user-created text prompt (individual word or short sentence) can be used to generate two images in the Image Generator module. We use a text-to-image network, the stable diffusion model \cite{stabilityai} with the default parameters to generate the first image and the Google Search API \cite{googleapi} to generate the second image. All generated images are also stored on the server. \\
Alternatively, different variations of an image can be created in the Image Augmentation Module. We rely on Pytorch's torchvision library to create four kinds of image augmentations. 

\paragraph{Failure Summarizer}
The Disaggregated Failure Metrics Generator takes all failure instances as input and calculates disaggregated failure metrics. The Visualizer presents failure instances and allows users to group, ungroup, delete them, and add interactive text strings to the zoomable canvas. This step helps users to find patterns amongst failure instances and to come up with failure-specific remedies. We used Fabric.js's Javascript HTML5 canvas library \cite{fabricjs} to implement the Visualizer.
\section{Expert Review}
We conducted a user study with UX experts to gather feedback on the usefulness of \system in exploring pre-trained AI models. Specifically, our objective was to (1) evaluate the failure exploration workflow of \system, (2) collect feedback on the  usefulness of our tool, and (3) evaluate the overall user experience of \system. 

We recruited 10 participants for the study, initially from our connections in the industry and then through snowball sampling, aiming to collect feedback from a skilled group of UX practitioners rather than a comprehensive sample. To avoid knowledge advantages, we recruited different participants for this study than for the formative interviews. For this study, we included UX practitioners with and without AI-specific experience. Our participants comprised eight UX designers and two product managers. Seven participants reported previous experience in the design of AI-powered applications. 

Each session lasted 60 minutes, and participants could opt-in to receive a small gift of university merchandise. 
Sessions were conducted individually (ten sessions in total) using the Zoom video conferencing system. At the start of each session, we introduced the participants to the application use case of our computer vision-based language learning app. Then, we asked the participants to imagine how they would design the user experience for the application. 
We asked participants to explore the capabilities and limitations of the DETR Object Detection~\cite{detrhugging} AI model and to derive UX design considerations based on their analysis. 
We also provided participants with a potential user profile \cite{holzinger2022} similar to Section~\ref{sec:ux} and three image samples, but participants could also create their own target groups and provide samples.
We set up the study as a comparative assessment. In one part, participants were first asked to perform the task using HuggingFace's interactive model card~\cite{detrhugging} for DETR, which includes model documentation and an API interface (playground) to test the model with image inputs. In the other part, we provided participants with a detailed walkthrough of \systems workflow and features. The participants then performed the same task using our tool. We randomly counterbalanced the order of the two parts to reduce order and carryover effect. Once the participants completed the task, they engaged in an open discussion with the study coordinator to report on the design considerations and their understanding of model limitations. They also provided feedback on the tools, and discussed how designers would use \system for their own AI UX work. Finally, at the end of the study, all participants completed a user experience questionnaire (UEQ-S)~\cite{ueq}. 

\subsection{Analysis}
We used a mixture of quantitative and qualitative methods to analyze the results. For the quantitative analysis, the first author watched all recordings and annotated video segments to quantify how much time participants spent on different views and the frequency and sequential order of feature usage, and calculated descriptive statistics. This was done for both \system and HuggingFace. To assess whether \system helped UX practitioners explore and analyze failure cases, we also extracted all outputs of the failure engine from the screen recording (failure mode tags) and all failure notes created (design synthesis canvas). We analyzed participants' design synthesis boards by clustering failure groups across participants. For the qualitative findings, we analyzed interview transcripts inductively corresponding to our evaluation objectives. 
 
\subsection{Findings}
Our findings include observations about \systems usage, usefulness, and user experience. We constantly compare participants' use of \system with HuggingFace's interactive model card, which is today's status quo for people to understand pre-trained models.

\subsubsection{User-centered Failure Exploration with \system}
 
 \begin{figure*}[]
\centering
\includegraphics[width=\textwidth]{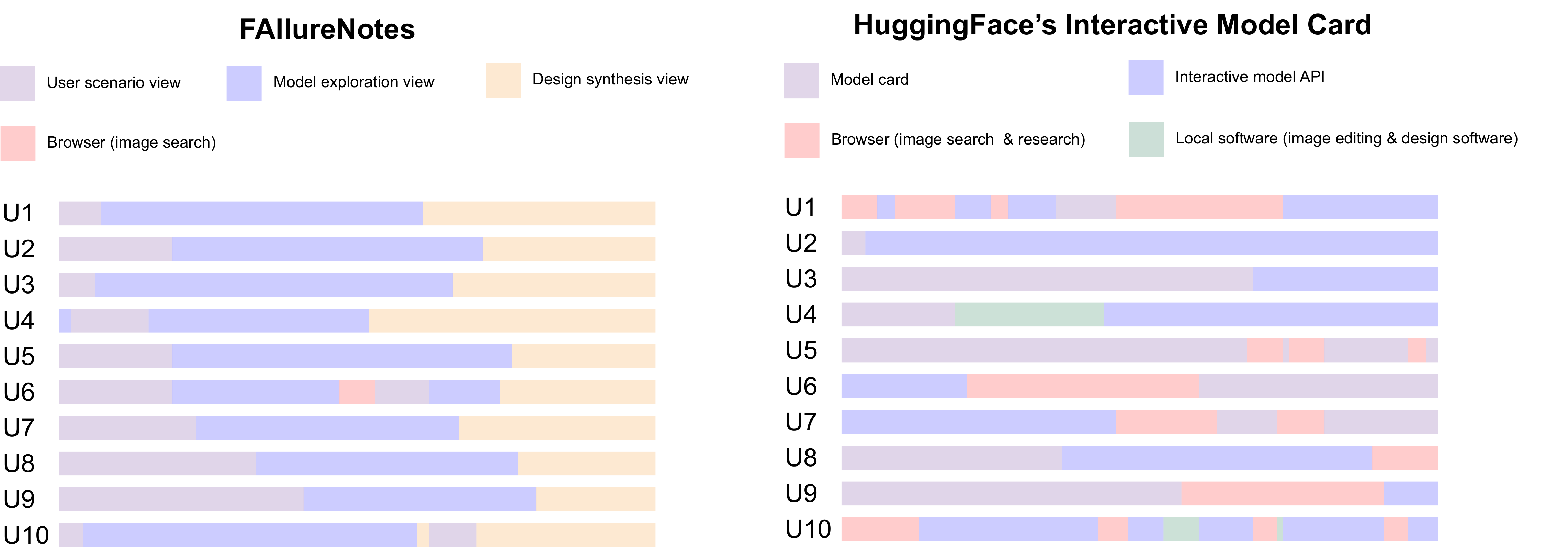}
\caption{Findings from Evaluation: Participants' workflow with \system (left) and HuggingFace (right). The bar's color indicates the view, and the length of the bar encodes time. }
\Description{The figure shows fAIlureNote's and HugingFace's system usage log data in comparison.  }
\label{figure:evaluation}
\end{figure*}

All participants followed a fairly linear process as they examined the DETR model' failures using \system (see Figure~\ref{figure:evaluation}). Only U6 and U10 went back to create additional user scenarios after examining the model for failures. Across all sessions, participants spent 19 percent of their time on the user scenario view, 49 percent on the model exploration view, 32 percent on the design synthesis view, and 1 percent on the browser. With \system, there was little need to switch between different tools. In contrast, on HuggingFace, designers left the interactive model card for 24 percent of their time to download images from Google, research information about the dataset, create data augmentations with local software, or sketch UI elements in Figma. 

In addition to the persona and scenario provided, designers had little difficulty imagining new user groups and scenarios. Across all sessions, participants created 0.9 users and 2.5 scenarios with \system. For example, U1 designed the fictional language learning app for ``Geena - a cook who wants to learn about food in Japan'', or U5 for ``Maria - a professional who moved to Madrid''.
Often the scenarios helped envision new images a user would submit to the AI. Most participants used the system's built-in image generation function to generate sample images for examination. The image generation function was used an average of three times per session, and only once did a product designer search Google for images (U6). With HuggingFace, six out of ten participants relied on Google to download images to test the model. 
Image augmentations helped designers challenge the AI model. Four out of ten designers used the image augmentation module of \system. During the HuggingFace session, U5 and U10 also created image augmentations using local image editing software. Most times, participants came up with their own prompts, while the prompt generation module was used only once by U7. 
 
 Most participants tested the AI model iteratively. Regarding the time spent, participants were slightly faster with HuggingFace's Model API (participants tested one image every 5 minutes and 2 seconds). With \system, on the other hand, it took 5 minutes and 59 seconds for each image tested. One reason is that \system provided expansive details and affordances to understand model failure. In addition to uploading an image and reviewing the model output (supported by both \system and HuggingFace), with our tool practitioners also need to annotate the image and review the failure engine's output.  
 
 All 10 participants derived design considerations (i.e., failure notes) with \system. Across all sessions, designers and product managers created 1.6 failure groups and 1.4 recovery strategies. In contrast, no designer was able to derive design considerations in written or visual form using the HuggingFace model API. In HuggingFace, five designers verbally discussed failure patterns (e.g., U10 noted that the AI only seems to detect objects in the foreground or that the model fails with rotated images) but had difficulty translating that into formal design considerations. By supporting designers to articulate different failure types, designers were able to ideate on design considerations to mitigate these failures. Further, discovering failure patterns requires looking across multiple failure cases, but designers cannot save and synthesize insights across all tested images with HuggingFace's interactive model card. 
 
In summary, while HuggingFace's model card allowed a slightly faster model exploration (but shallower understanding), participants often switched between different tools. Most notably, \system significantly outperformed HuggingFace with respect to the quality of model assessment and design considerations.

\subsubsection{Perceived Usefulness}
\begin{figure*}[h]
\centering
\includegraphics[width=\textwidth]{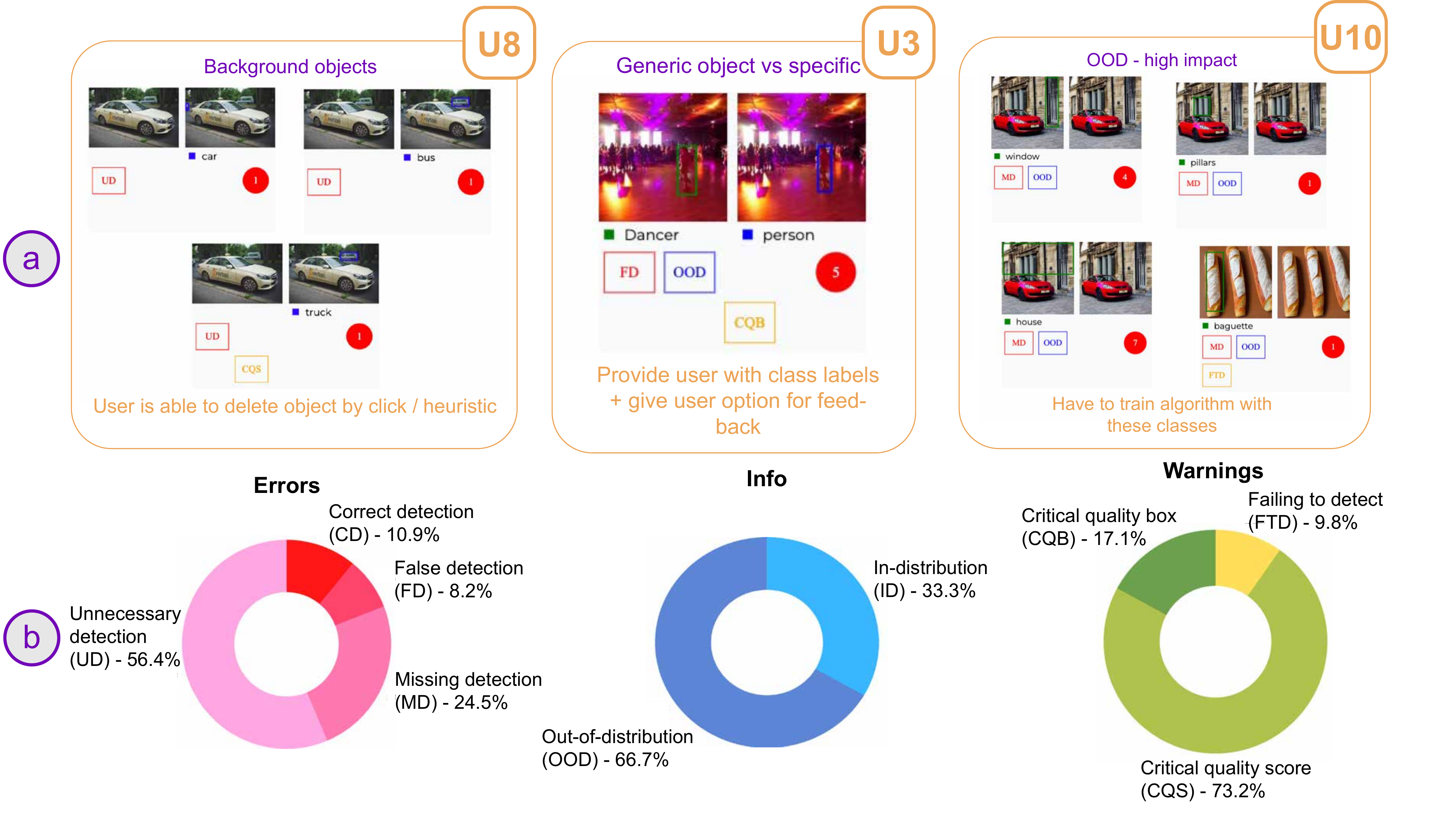}
\caption{Findings from Evaluation: (a) Example failure notes of three participants and (b) an aggregated overview of the error, info, and warnings tags participants experienced. }
\Description{The figure is divided into two parts. The first part shows three design synthesis boards participants created during their failure exploration session. The second part shows on aggregated overview of the error, info and warnings tags participants experienced. }
\label{figure:synthesis}
\end{figure*}
When asked during the post-task open discussions, designers with prior AI design experience design and knowledge of HAI guidelines (n=7) saw value in \system. Reflecting on the two tools used in our study, U5 said \say{If I'd be the product designer and [the AI] is really a fundamental part of this product, I would use \system instead of the HuggingFace version}. U6 said \say{It’s one thing to know the [HAI] guidelines and another to apply them in your work, and the tool made it very explicit how the user interface should respond in case of an error}. Participants found that \system gave them a more structured approach to the task of model behavior analysis than today's interactive model cards. 
In particular, designers appreciated the user scenarios embedded within the model exploration context. For example, U1 commented, \say{I liked the fact that you could go and swap through the different personas and work on the different storylines. I think that is useful for setting the scope}. 

Further, four participants explicitly remarked on the design synthesis view, allowing them to group failure instances into different categories. \say{I like having the option of grouping failures and having this canvas to do this. It is easier to connect stuff in your head with this visual representation, where you can connect different pictures or elements together} (U4). U7 and U10 said that such a feature would have helped them enormously during their work. U7 said that \say{a broad overview of possible failures would have given us a better way to organize our communication with the user}. She would have used the failure canvas as a boundary object~\cite{star1989structure}: \say{there were always discussions between ML and product, and it would be really nice to communicate [the failure groups] visually to the whole team rather than writing it in words}.

Figure~\ref{figure:synthesis}a shows some of the failure groups and recovery mechanisms participants created during their session with \system. U8 and U10 experienced a group of failure cases where the model detected background objects. As a remedy, U8 imagined that a user could delete unwanted objects on click within the language learning application, or work out a heuristic with an engineer that would prevent showing background objects in the first place. Three participants (U3, U4, and U7) found a group of failure cases where the AI model would predict objects with a higher level of abstraction (e.g., detecting a person instead of a dancer). U3 added a failure group named ``Generic vs. specific object'' and imagined providing users with an explanation of what classes the model can predict while also giving users a way to provide feedback. Lastly, five participants (U1, U2, U6, U8, and U10) discovered a group of objects that the user expected the model to detect, but the labels were out-of-distribution. U10 named this group ``OOD - high impact'' and noted that the model would need to be retrained by engineers. 

Across all participants, \system's failure engine helped identify several failure modes (see Figure~\ref{figure:synthesis}b). The most common error type was unnecessary detections (56.4\%),  followed by missing detections (24.5\%) and false detections (8.2\%). For 66.7\% of objects, the annotated ground truth was outside of the model's capabilities (OOD) and 33.3\% of annotated objects were in-distribution (ID). The failure engine's most commonly displayed warning was the critical quality of score (CQS) with 73.2\%, followed by critical quality of box (CQB). In 9.8\% of cases, the AI model did not detect any object.

\subsubsection{User Experience}
Overall, designers found \systems user interface supportive and relatively intuitive to use. U1 contrasted HuggingFace's model card to \system and said, \say{The UI of your tool was a lot friendlier for designers, especially if you are not familiar with code. It's less technical, not daunting, and is frictionless}. U6 commented on \system saving him valuable time during his manual image search and failure exploration. \say{What was also cool when I tried [HuggingFace], and I stumbled upon the cooking pot\ldots I was thinking maybe it was the reflection, maybe this and that, and [\syste] would have probably just given me it is out-of-distribution. So it would have saved me a lot of headaches to search for more cooking pots with less reflection, better lighting, or whatever}.

However, the failure modes were difficult for designers with little prior experience with object detection. For example, U1 had difficulty understanding automated matching and was confused about the coloring between user expectation and model prediction. She also commented that it takes time to become familiar with the acronyms (e.g., FD for false detection). Sometimes, designers wanted to provide feedback on the failure engine's classification. U8 said: \say{It would be nice to say, this was actually not an error and disagree with the machine}. He also suggested a different user experience from the canvas to group failure cases and said: \say{I wonder if this kind of interface is the most efficient for grouping\ldots I imagine you get like hundreds of these failure cases}. Instead, he suggested a UX that allows designers to multi-select failure cases and create failure groups by dragging items into folders. We plan to incorporate these suggestions in future iterations of \system and explore ways to revise and embed more designer-centered interactive model cards into \system.

Based on the usability questionnaire, on a seven-point scale, participants rated \system to be supportive rather than obstructive (\textit{mean}=5.91, \textit{SD}=0.67) and efficient rather than inefficient (\textit{mean}=5.00, \textit{SD}=1.12). Participants rated our tool as exciting rather than boring (\textit{mean}=5.10, \textit{SD}=1.04) and interesting rather than not interesting (\textit{mean}=5.8, \textit{SD}=0.60). They evaluated it as easy rather than complicated (\textit{mean}=4.70, \textit{SD}=1.10) and clear rather than confusing (\textit{mean}=4.40, \textit{SD}=1.36). Lastly, designers found our tool inventive rather than conventional (\textit{mean}=5.60, \textit{SD}=0.80) and leading edge rather than usual  (\textit{mean}=5.60, \textit{SD}=0.80). Encouraged by the positive feedback, we plan to run follow-up deployment studies in real-world AI design tasks with improved onboarding experience (without having the guided walkthrough). 
\section{Discussion and Future Work}
In this section, we discuss the utility of our developed system, \systems generalizability, and the underlying assumptions and limitations of our two studies.

\subsection{Utility of \system}
AI, often seen as a ``general-purpose'' technology that can be incorporated into numerous applications, carries enormous potential and risks. AI failures can have negative consequences and create physical, psychological, or financial harm to humans~\cite{mcgregor2020, gebru2021}. Yet, designers and engineers lack the means to foresee problems with AI models in real-world use~\cite{hong2021}. As prior literature has shown, human-centered AI largely follows an ``AI-first''workflow~\cite{subramonyam2022a}. AI engineering practices have established error analysis tools~\cite{wu2019, eyuboglu2022}, processes, and artifacts for transparent reporting~\cite{mitchell2019, anamaria2022}. However, it is difficult to anticipate downstream application performance and use through upstream model documentation. To facilitate this transition, it has been proposed that designers must acquire a ``designerly understanding''~\cite{yang2020} of AI models as a necessary first step to AI application design. However, current design processes and tools fail to meet this objective. Current HCI research has contributed guidelines \cite{pair2019, amershi2019} and tools for designing AI experiences \cite{subramonyam2021, hong2021, jansen2022}. With \system, we have demonstrated an approach for early failure discovery of pre-trained object detection networks. Using \system, UX professionals can interactively explore AI models using insights from their user research. Our evaluation and the participants' feedback offer evidence that our system bridges understanding pre-trained models and designing application experiences around them.

Further, since our system is intended to support the incorporation of user research data, we can imagine designers collaboratively working with potential end-users or domain experts to explore, categorize and prioritize different failures. Oftentimes, designers alone cannot assess whether the model behavior constitutes a failure as it can be subjective to use contexts. As opposed to current \textit{Wizard of Oz} testing, \system can provide end-users with a more realistic expectation of AI behavior and performance while informing design considerations. In designing \system, we aimed to support designers along the entire model behavior analysis journey---from generating samples, reviewing model outputs, and disaggregated model assessment---all within the same tool. As the results of our evaluation show, our image generation, editing, and prompting support features allow designers to stay within the system's workspace without switching application context. Given the range of artifacts and tools designers need to navigate in AI experience design, an integrated tool and workflow such as \system can greatly reduce friction in design work. 

However, we did not implement the transition between failure exploration and UI/UX design or modeling changes in our tool. To achieve a \textit{failure-driven design process}, we aim to explore extensions that also allow designers to sketch, design, or prototype AI-powered applications (a relevant example is ProtoAI \cite{subramonyam2021}). Additionally, our workflow and functionalities (e.g., image generation module) work particularly well for early design phases when data is scarce. However, model testing is an ongoing challenge due to potential changes in user behavior or distribution drifts. Building on model behavior analysis tools for AI engineering (e.g. Affinity \cite{cabrera2022}), we hope to see the HCI community develop similar tools that are specifically targeted at product designers or product managers. This could include functionalities such as uploading existing datasets, slicing data into subsets \cite{polyzotis2019}, or filtering instances \cite{cabrera2022}. 

\subsection{Generalizability}
Our current prototype primarily supports the exploration and evaluation of object detection models (single-stage \cite{girshick2013, shaoqing2015, girshick2015} and two-stage detectors \cite{redmon2015, lin2017, carion2020}). In order for the system to be used for other machine learning tasks, the ground truth annotator, AI prediction module, failure engine, and prompt generator would need to be adapted. However, we believe that our approach of using generative models to augment data for exploring model failures is scalable to many other machine learning tasks, such as image classification or semantic segmentation in computer vision and text classification in natural language processing. For example, future studies could use large language models \cite{brown2020} to generate text inputs that could be annotated before being fed into text classification networks. However, if the model to be explored is a generative model, \system's functionalities and workflow most likely would have to change (e.g., the annotation and failure modes would be different). Future HCI work could explore how to support failure exploration for large pre-trained generative models (``foundation models'') such as image generation models \cite{stabilityai} or large language models \cite{percy2022}. 

Moreover, our tool primarily focused on failures at the observation layer of computer vision systems (see Table \ref{table:failure_modes}: A taxonomy of failure modes in computer vision). Others could explore how tools can support UX practitioners in understanding errors that occur at the sensing and reaction layers. Assessing input and label quality is an important part of the error analysis process \cite{cabrera2021} that our work has not addressed in detail. We also do not cover complex systems where the system contains multiple machine learning models. Previous work \cite{nushi2018} has highlighted these challenges from a technical perspective, but future work may help non-technical users (e.g., product managers) in (a) understanding the ML pipeline and (b) discovering failures at different system levels.

\subsection{Assumptions and Limitations}
Our approach primarily emphasizes failure cases involving end-user experiences, but the scope of AI failures can be broader \cite{mcgregor2020}. In cases such as adversarial attacks \cite{huang2017, naveed2018}, or privacy breaches \cite{xiong2021}, it is crucial to include a wider range of stakeholders. Additionally, \system also does not directly consider systematic issues related to fairness, accessibility, or transparency \cite{madaio2020}, such as supporting testing model performance for different genders or races \cite{buolamwini2018}. Future work should examine how early-stage model probing could push designers toward sociotechnical definitions of AI technology failure.

Our evaluation study demonstrates the shortcomings of interactive model cards \cite{mitchell2019} in assessing the contextual fit of pre-trained AI models. In essence, none of our participants was able to find aggregated failure patterns or derive UX design considerations for failure recovery with HuggingFace's model card. However, our study did not cover other aspects of design with AI models (creating UIs) and the time given to participants was limited. We also acknowledge that the number of participants (n=10) is small. We also did not test our system in a real-world work setting nor did we include AI engineers (e.g., to assess the failure groups designers found). We are interested in how failure exploration and analysis tools such as \system can support product teams collaboratively in a real-world setting. Future work can focus on the design-engineering boundary \cite{subramonyam2021towards} and support a collaborative error analysis process.

\section{Conclusion}
Artificial Intelligence (AI) is becoming an integrated part of our society and lives. However, despite AI's promise to enable novel user experiences and services, errors will remain an inevitable byproduct of AI-powered applications. To proactively anticipate and address AI failures, UX designers need access to the underlying technology to understand the model's capabilities and limitations. In this work, we introduced a \textit{failure-driven design} approach to AI, a workflow that encourages designers to explore model behavior and failure patterns early in the design process. Our implementation of \system, a designer-centered failure exploration and analysis tool, supports designers in evaluating models and identifying failures across diverse user groups and scenarios. We demonstrate how \system can support designers in operationalizing HAI guidelines and provide users with a path forward from AI failures.

\begin{acks}
We thank our reviewers and study participants for their time and feedback. We also thank Dipti Ganeriwala and Raisul Ahsan for providing feedback on early prototypes and Natalija Wollny for her help with the demo video.
\end{acks}
 \balance
 
\bibliography{99_refs}
\pagebreak

\appendix

\section{Appendix}

\begin{figure*}[b]
\centering
\includegraphics[width=0.9\textwidth]{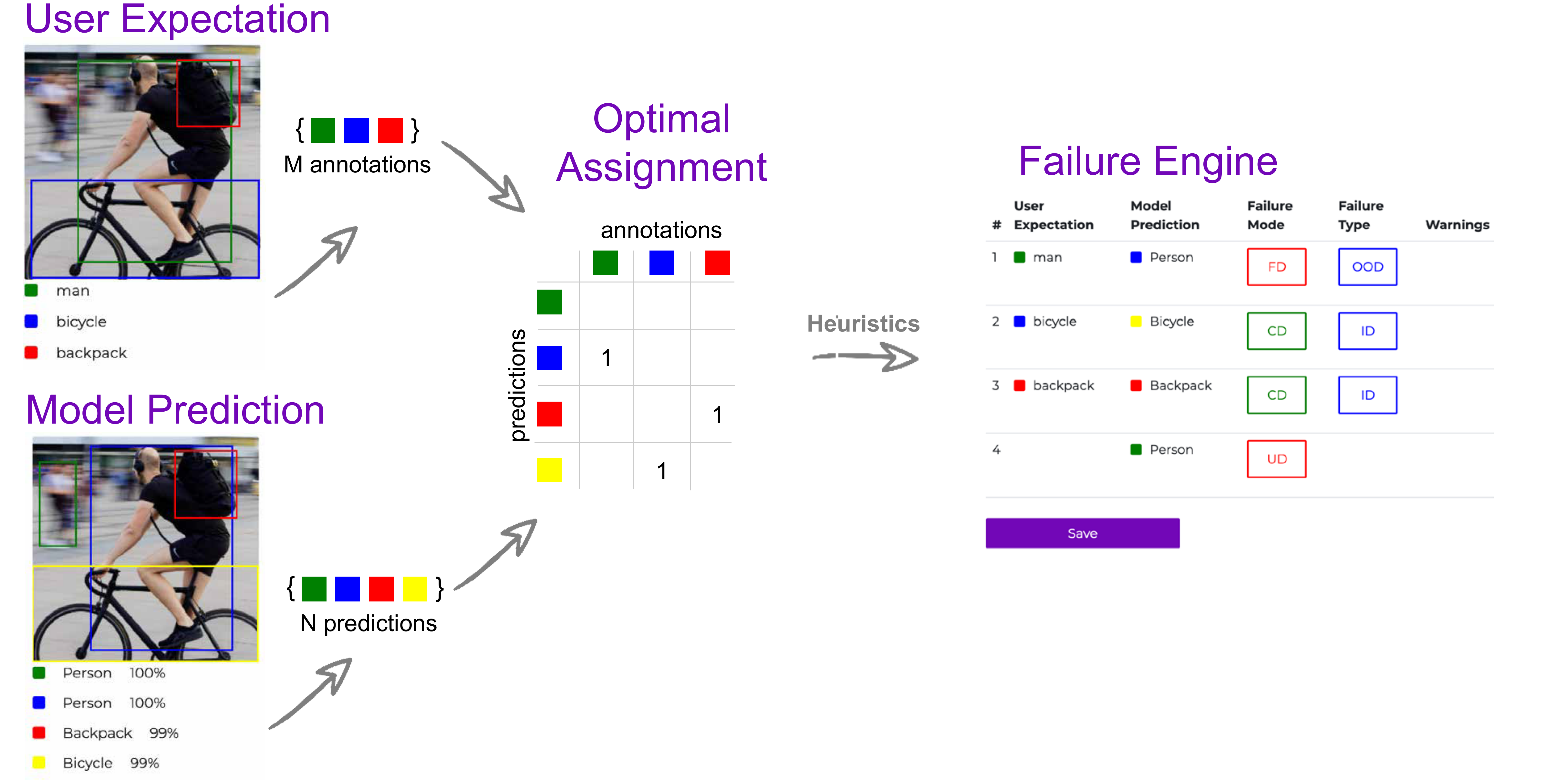}
\caption{\systems Failure Engine: A matching algorithm finds an optimal assignment between M annotations and N predictions. }
\Description{The figure illustrates the system's failure engine. It shows how M annotated and N predicted objects are optimally assigned with a matching algorithm. For this, the figure shows a matrix (M x N). The failure engine then differentiates different failure modes based on the optimal assignment and simple heuristics. }
\label{figure:engine}
\end{figure*}
\subsection{Technical Details}
Here we outline the technical details of \systems failure engine which is inspired by the DETR loss~\cite{carion2020}.

Figure~\ref{figure:engine} illustrates the matching problem. Each image has M annotated objects and N predicted objects. In our example, the user expected the AI model to detect three objects (a man, a backpack, and a bicycle). However, the AI model predicted four objects (a person, another person, a backpack, and a bicycle). The goal is to match the annotations and model predictions based on a matching cost. 
More formally, let us denote \(y\) for the annotated set of objects and \(\hat{y}\) for the predicted set of objects. 
To find a matching between these two sets (annotations \& predictions) we search for a permutation \(\sigma \in \Sigma\) with the lowest costs:
\[\hat{\sigma} =  \underset{\sigma \in \Sigma}{\arg\min} \sum_{i}^{N} L_{match}(y_i, \hat{y}_{\sigma}(i)) \]

The \textit{optimal assignment} is essentially a \(MxN\) matrix where each match is indicated with ``1''. It is calculated with the Hungarian algorithm \cite{stewart2015}. The algorithm takes a \textit{matching cost} as an input and returns the optimal assignment between annotated and predicted objects.
Formally, \(L_{match}\) is a pair-wise matching cost between ground truth \(y_i\) and prediction \(\hat{y}\) with index \(\sigma (i)\).  It returns the optimal assignment \(\hat{\sigma}\) between the annotated and predicted objects. Each object consists of a class label and a bounding box. For instance,  each annotation can be seen as \(y_i=(c_i, b_i)\) where \(c_i\) is the class and \(b_i\ \in \mathbb{R}^4 \) is a vector that defines the four points of a bounding box.
The matching cost needs to take both the class labels and bounding boxes into account. We can provide two examples to illustrate this point. The AI model predicted two ``person'', one on the bicycle and another in the background. By reviewing the user's expectations and model prediction we can see that the ``person'' on the bicycle (and not in the background) should be matched to the annotated ``man''. Alternatively, we can imagine a scenario where the user's bounding boxes do not overlap strongly with the model prediction but he or she still meant the same object. Formally, we can define the \textit{matching cost} as:

\[L_{match} = \gamma_{class} * L_{class}(c_i, \hat{c}_{\sigma(i)}) + \gamma_{box} * L_{box}(b_i, \hat{b}_{\sigma(i)})\]
The matching cost consists of a \textit{class loss} and a \textit{bouding box loss} weighted by two hyperparameters. We set both hyperparameters \(\gamma_{class}\) and \(\gamma_{box}\) to 0.5. The \textit{class loss} compares the class labels of all annotations and predictions. We would like the cost to be low if the classes match (i.e, backpack, bicycle) and high in case they are different. We defined the \textit{class loss} as:
\[
{L_{class(c_i, \hat{c}_{\sigma(i)})} = \begin{cases} 
      0 & c_i = \hat{c}_{\sigma(i)} \\
      1 & else
   \end{cases}}
\]
In other words, the cost is set to zero if the class labels of the annotation and prediction match and one otherwise. In order to assess whether two objects are matching we would also like to take the bounding boxes into consideration. The \textit{bounding box loss} is a linear combination of the \(l1\) loss and the generalized IoU loss \cite{rezatofighi2019}.
\[ L_{box}(b_i, \hat{b}_{\sigma(i)}) = \lambda_{l1} * \|{b_i - \hat{b}_{{\sigma}(i)}}\|  + \lambda_{iou} * L_{iou} \]
\\ 
where \(\lambda_{l1}\) and \(\lambda_{iou}\) are hyperparameters (we set both to 0.5 again). The \(l1\) loss calculates the absolute distance between each point of the bounding box. The generalized IoU loss helps to make the loss scale invariant.
Once the Hungarian algorithm returns  the optimal assignment we can use simple heuristics to classify different failure modes.

\end{document}